\documentclass[pdflatex,sn-nature]{sn-jnl}

\usepackage{graphicx}
\usepackage{amsmath,amssymb,amsfonts}
\usepackage{booktabs}
\usepackage{microtype}
\usepackage{hyperref}

\begin{document}

\title[Causal Discovery in the Era of Agents]{Causal Discovery in the Era of Agents}

\author[1]{\fnm{Yujia} \sur{Zheng}}\equalcont{These authors contributed equally to this work.}

\author[1]{\fnm{Vishal} \sur{Verma}}\equalcont{These authors contributed equally to this work.}

\author[1]{\fnm{Mantej} \sur{Gill}}\equalcont{These authors contributed equally to this work.}

\author[1]{\fnm{Haoyue} \sur{Dai}}

\author[1]{\fnm{Peter} \sur{Spirtes}}

\author[1,2]{\fnm{Kun} \sur{Zhang}}

\affil[1]{\orgname{Carnegie Mellon University}}
\affil[2]{\orgname{Mohamed bin Zayed University of Artificial Intelligence}}

\abstract{Recent attempts to combine large language models (LLMs) with causal discovery ask models to infer pairwise directions, propose graph structures, or inject language-model outputs as priors and constraints. These approaches promise faster analysis, but they also obscure whether a causal evidence is supported by data and assumptions or by textual associations, prompt artifacts and hallucinated mechanisms. We argue for a different role for agents in causal discovery. Agents should inspect data, retrieve context, explain method assumptions and clarify graph outputs, but they should not supply edges, orientations, priors, constraints or causal conclusions. We propose the principle that agents assist the workflow, while causal claims remain grounded in data, explicit assumptions, formal algorithms, diagnostics and user or domain-expert decisions. We instantiate this principle in \textit{causal-learn+}, an online platform that coordinates data analysis, preprocessing, method recommendation, expert-knowledge incorporation, formal discovery and interpretation around the algorithmic ecosystem of \textit{causal-learn}. A case study on Big Five personality data illustrates agent-assisted pipeline of causal discovery without turning language-model unreliability into causal evidence. The platform is available at \href{https://causallearn.com}{causallearn.com}.}


\maketitle

\section{Introduction}

Causal discovery aims to learn causal relations from observational data \citep{spirtes2000causation}. This matters because interventional experiments are often expensive, slow, unethical or impossible. In genomics, ecology, neuroscience, epidemiology and many other scientific domains, large observational datasets are often easier to obtain than carefully controlled interventions \citep{glymour2019review}. In these settings, causal discovery can turn data into structured causal hypotheses that scientists can inspect, challenge and refine.

The field has developed several major approaches to this problem. Constraint-based methods, such as PC \citep{spirtes2000causation} and FCI \citep{spirtes1995causal}, use conditional independences to restrict the space of possible graphs. Score-based methods, such as GES \citep{chickering2002optimal}, search for graphs that optimize a statistical criterion. Functional causal model approaches exploit asymmetries in noise distributions, functional form or distribution to orient relations that conditional independences alone cannot distinguish \citep{shimizu2006linear,hoyer2008nonlinear,zhang2009identifiability}. Latent-variable causal discovery methods, including recent work based on generalized independent noise and rank constraints, address the fact that many scientific variables are hidden, measured indirectly or confounded \citep{xie2020generalized, dong2024versatile}. These families differ in the assumptions they make and in the conclusions they support. Stronger assumptions can yield more informative causal conclusions, but no method is assumption-free.

This dependence on assumptions makes causal discovery difficult for practitioners, especially those without specialized training in causal methodology. A user must understand the scientific context, decide which variables belong in the analysis, reason about data nature (e.g., missingness), choose among algorithm families, interpret partially identified graph objects and avoid reading more causality into the output than the assumptions justify. Existing software libraries have made many algorithms available, but availability alone does not make them easy to use well. The practical bottleneck often lies between domain expertise, data analysis and causal methodology.

Recent work has begun to introduce large language models into causal discovery. Some methods query a model for pairwise causal directions or full graphs from variable names and natural-language descriptions \citep{long2023can,kiciman2023causal}. Others incorporate model outputs as priors, constraints or graph refinements inside statistical discovery procedures \citep{long2023causal,darvariu2024large,takayama2024integrating}. These directions aim to make causal discovery more direct and efficient by exploiting the broad knowledge encoded in pretrained models.

However, directly injecting language model outputs into graph construction creates a basic problem. Causal discovery seeks evidence about causal relations, whereas a language model is trained to predict patterns in text. If language model outputs enter the search procedure, a discovered relation may reflect statistical evidence in the dataset, background knowledge from the pretraining corpus, prompt wording, common scientific belief or a hallucinated explanation. These sources are difficult to separate after the graph has been produced. This ambiguity is especially problematic for causal discovery, where the value of a conclusion depends on knowing which assumptions, data and algorithms justify it. Therefore, we take a conservative position:

\begin{center}
\textit{Agents should assist causal discovery, not replace it.}
\end{center}

Rather than allowing LLMs to participate directly in graph discovery, replace search modules or act as conclusion providers, we use agents to support the surrounding workflow. They can coordinate tools, retrieve methodological and domain knowledge, invoke modular skills, and smooth the analysis pipeline across different stages of causal learning. In this role, agents guide practitioners through data understanding, assumption clarification, method selection, execution and interpretation, while also providing domain-specific context when needed. Crucially, these agentic activities are embedded within a rigorous causal analysis pipeline. They improve usability and coordination without introducing LLM hallucinations as causal evidence, and users retain control over the key decisions that affect the final conclusion.

We develop this principle through \textit{causal-learn+}, an online agentic system for causal discovery around the algorithmic ecosystem of \textit{causal-learn} \citep{zheng2024causal}, an open-source platform that has gained much attention. The system treats agents as assistants to causal learning rather than replacements for discovery algorithms. Building on the principle that causal conclusions must be derived under explicit and appropriate assumptions, \textit{causal-learn+} organizes the analysis into traceable steps, including data inspection, assumption elicitation, method recommendation, tool coordination, algorithm execution, graph inspection and result interpretation. Agents support these steps by combining language understanding, retrieval over causal methods, domain knowledge, modular skills and external tools, while causal claims remain tied to the selected assumptions, executed algorithms, statistical diagnostics and user-approved decisions. In this way, \textit{causal-learn+} broadens access to causal discovery while preserving the evidential discipline required for reliable causal analysis.

\section{Agent Cannot Provide Causal Evidence}

The central risk in agentic causal discovery is not that agents are useless. It is that useful agent outputs can be mistaken for causal evidence. A language model can summarize literature, suggest possible temporal-order assumptions for user review, recognize common mechanisms and produce coherent explanations. These outputs can guide a user, but they do not constitute data-based causal evidence.

Causal discovery concerns which causal conclusions are supported by data, assumptions and formal procedures. If an agent proposes edges, orients relations, adds priors, chooses constraints or tunes thresholds, the resulting causal graph is no longer the output of a well-specified discovery method. A relation may reflect the dataset, the algorithmic assumptions, common beliefs in the pretraining corpus, prompt wording or hallucination. Once these sources mix, the graph's evidential status becomes ambiguous.

% This position differs from recent work that treats language models as imperfect experts, priors or statistical-discovery aids \citep{long2023causal,darvariu2024large,takayama2024integrating}. Such outputs may be useful, but their role should be external to causal evidence unless a user explicitly adopts them as assumptions. In that case, they should be recorded as domain input, not silently merged with the data-driven discovery procedure.

This matters because causal discovery often produces partially identified objects that are easy to overread. A CPDAG (Completed Partially Directed Acyclic Graph) is not a single DAG; it represents a Markov equivalence class of DAGs. A PAG (Partial Ancestral Graph) is not a fully observed causal model; it represents a Markov equivalence class of causal models that may contain latent confounders and selection variables. A latent node is not automatically a named scientific construct; and a Granger relation is not automatically an intervention claim. A fluent agent explanation can make such outputs sound more definitive than the method warrants.

The same risk appears in small workflow decisions. An agent might silently drop variables, standardize data, add a literature-based forbidden edge, convert temporal hints into required tiers, lower a rank threshold until a latent variable appears or orient an ambiguous edge for readability. These actions may look helpful, but they change what the algorithm treats as input or output. They should be visible user decisions, not hidden agent behavior.

The principle is simple: agents may support causal discovery, but they cannot provide causal evidence. Evidence for graph structure should come from observed data, explicit assumptions, formal algorithms, diagnostics and user or domain-expert decisions. In causal discovery, provenance is not bookkeeping; it is part of what a graph mark means. Agent outputs should therefore remain context, guidance or explanation unless explicitly adopted as assumptions or interpretations outside the formal core.

\section{How Agents Should Help Causal Discovery}

If agents should not provide causal evidence, their role must be defined positively. The right role is assistance around causal discovery: making the workflow easier to execute, understand and document, while leaving causal conclusions tied to formal algorithms and scientific judgment.

This role should satisfy five design criteria, consistent with broader calls for transparent agentic scientific systems \citep{multiagent2026transparency,xin2025towards}. Separation keeps agent suggestions outside the formal discovery core. Visibility makes preprocessing, assumptions and method choices explicit. Traceability links each graph output to data, parameters, diagnostics and user decisions. Reversibility lets users inspect or undo agent-supported steps. User approval prevents contextual knowledge from silently becoming a causal constraint.

First, agents can help users understand the data before discovery begins. They can summarize variables, detect missingness, flag unusual distributions, identify possible identifiers or time stamps and call visualization or statistical tools. These routine checks help users notice problems before they affect method choice or interpretation.

Second, agents can support preprocessing without silently making preprocessing decisions. Scaling, discretization, missing-data handling, variable exclusion and item recoding can all affect a discovered graph. An agent can explain alternatives, run checks and propose transformations, but consequential changes should remain visible user decisions. The assistant should make preprocessing deliberate rather than hidden.

Third, agents can guide algorithm selection. Causal discovery includes constraint-based, score-based, functional, latent-variable and time-series approaches. Users often know the scientific question but not the assumptions or graph objects associated with each algorithm. An agent can translate between scientific and methodological language by helping users consider whether latent confounding, temporal order, linear non-Gaussian structure or rank constraints are relevant.

Fourth, agents can incorporate expert knowledge as context. This may come from retrieval over documentation and papers, domain-specific resources or the model's own background knowledge. Such knowledge can explain measurements, suggest assumptions to consider or warn against implausible interpretations. It should not directly become an edge, orientation or constraint inside the discovery algorithm. Users or domain experts should explicitly approve any role it plays in the analysis.

Fifth, agents can coordinate tools. They can call data-analysis routines, launch selected algorithms, collect diagnostics and organize run outputs for user comparison. Tool coordination is different from causal judgment. The agent may execute a selected algorithm, but it should not alter the conditional-independence tests, scores, rank constraints, graph search, orientation rules or graph files that define formal discovery.

Finally, agents can help interpret and communicate results. They can explain graph notation, summarize assumptions, connect results to domain vocabulary and draft reports. This is valuable because causal discovery often returns partially identified graph objects that are easy to overread. The assistant should make those limits clearer, not convert an assumption-conditioned graph into a causal fact.

These principles motivate a design in which agents surround causal discovery rather than enter its inferential core. A useful agentic system should expose agent suggestions, user choices, algorithm settings and graph outputs as distinct objects. It should make causal discovery more convenient, transparent and accessible while preserving the distinction between assistance and evidence.

\section{\textit{causal-learn+}: Agentic Causal Discovery with Trust}

\begin{figure}[t]
    \centering
    \includegraphics[width=\linewidth]{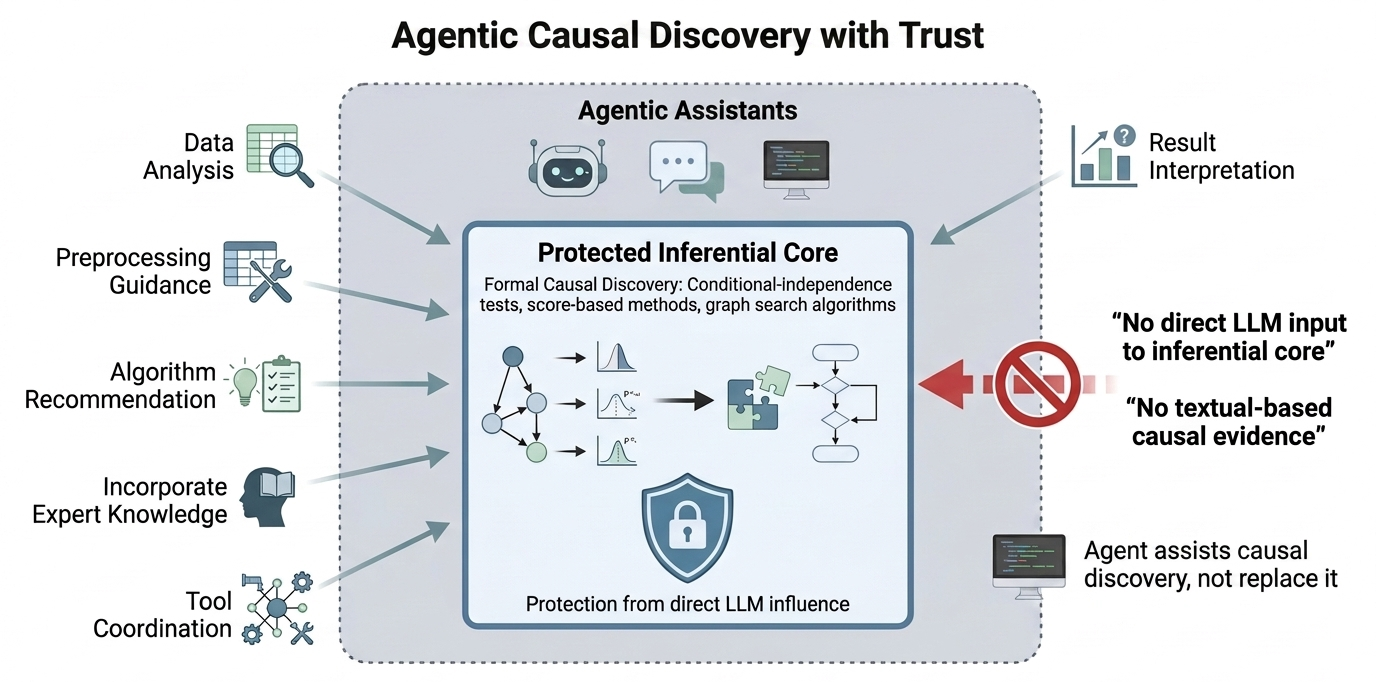}
    \caption{Framework of \textit{causal-learn+}. Agents support data analysis, preprocessing, algorithm recommendation, expert-knowledge incorporation and interpretation, while formal causal discovery remains tied to provably-correct algorithms, explicit assumptions and user-approved choices.}
    \label{fig:causal-learn-plus-framework}
\end{figure}

\textit{causal-learn+} instantiates these principles as an online agentic environment for causal discovery. As highlighted, we do not allow agents to infer causal relations. It lets agents coordinate the workflow around a protected algorithmic core. The user moves through data analysis, preprocessing, algorithm recommendation, expert-knowledge incorporation, formal causal discovery and interpretation, while the discovery step itself is performed by algorithms from \textit{causal-learn}. The design goal is guided access without changing the evidential semantics of the underlying methods. The platform (Figure \ref{fig:causal-learn-plus-framework}) therefore records agent suggestions, user choices, algorithm parameters, diagnostics and graph outputs as distinct parts of the analysis.

The workflow begins with data analysis. After upload, agents can call tools to summarize variables, missingness, data types, sample size, distributions and simple relationships. This stage helps the user judge whether the dataset is ready for causal discovery and which variables require attention.

The next stage is data preprocessing. Agents may suggest scaling, missing-data handling, variable exclusion or recoding, and explain how each choice can affect the downstream graph. These suggestions are not causal evidence. They become part of the analysis only when the user accepts them as preprocessing decisions.

Algorithm recommendation comes next. Based on the data profile and user question, agents can present candidate method families, their assumptions and relevant diagnostics for user selection: constraint-based methods when conditional independences are central, score-based methods when a scoring model is appropriate, functional methods when stronger noise or functional assumptions are plausible, latent-variable methods when measured variables may indicate hidden structure, and time-series methods when temporal prediction is the target.

Expert knowledge can be shown alongside the workflow as context through retrieval-augmented generation, documentation search, domain references or the model's own background knowledge. This information can explain variables, assumptions and plausible interpretations. It is not automatically a graph prior. Any domain constraint must be explicitly approved by the user or domain expert.

Formal causal discovery is the protected core. Conditional-independence tests, scores, rank constraints, graph search, orientation rules and graph outputs are computed by explicit algorithms. No LLM or agent participates in these calculations. This is where algorithm-indexed graph evidence is produced, and it remains tied to the selected data, assumptions, parameters and diagnostics.

After the run, agents help with interpretation. They can explain graph marks, summarize diagnostics, organize method outputs and draft reports that state the assumptions behind each graph-level statement or user interpretation. Interpretation is especially useful for users who are new to CPDAGs, PAGs, latent-variable outputs or Granger relations. The agent's role is to make the result understandable without changing what the algorithmic output licenses.

\begin{table}[t]
\centering
\caption{Causal discovery methods in \textit{causal-learn+}.}
\label{tab:causal-learn-methods}
\footnotesize
\setlength{\tabcolsep}{4pt}
\renewcommand{\arraystretch}{1.0}
\begin{tabular}{@{}p{0.18\linewidth}p{0.76\linewidth}@{}}
\toprule
\textbf{Category} & \textbf{Methods} \\
\midrule
Constraint-based & PC \citep{spirtes2000causation}, MV-PC \citep{tu2019causal}, FCI \citep{spirtes1995causal}, CD-NOD \citep{huang2020causal} \\
\midrule
Score-based & GES \citep{chickering2002optimal}, DGES \citep{li2024causal}, A$^\ast$ \citep{yuan2013learning}, Dynamic Programming \citep{silander2006simple}, GRaSP \citep{lam2022greedy}, BOSS \citep{andrews2023fast} \\
\midrule
Function-based & ANM \citep{hoyer2008nonlinear}, PNL \citep{zhang2009identifiability}, LiNGAM \citep{shimizu2006linear}, DirectLiNGAM \citep{shimizu2011directlingam}, VAR-LiNGAM \citep{hyvarinen2010estimation}, RCD \citep{maeda2020rcd}, CAM-UV \citep{maeda2021causal} \\
\midrule
Latent variables & GIN \citep{xie2020generalized}, RLCD \citep{dong2024versatile} \\
\midrule
Granger causality & Linear Granger causality \citep{granger1969investigating,granger1980testing,shojaie2010discovering} \\
\midrule
CI tests & Fisher's $z$ test \citep{fisher1921014}, missing-value Fisher's $z$ test, Chi-square test, KCI/kernel tests \citep{zhang2011kernel}, G-square test \citep{tsamardinos2006max} \\
\midrule
Scores & BIC \citep{schwarz1978estimating}, BDeu \citep{buntine1991theory}, generalized score \citep{huang2018generalized} \\
\bottomrule
\end{tabular}
\end{table}

On top of the workflow components discussed above, the online implementation provides an additional practical advantage. Users can conduct the workflow at \textit{causallearn.com} without installing packages, configuring local computation or writing code for every step. This lowers the threshold for causal discovery while preserving the central requirement of the paper: agents may make the process easier to use, but causal discovery remains grounded in formal algorithms and explicit user decisions.

\section{Real-world Example}

We present a real-world example of causal discovery with personality questionnaires, a familiar setting that is surprisingly difficult for causal analysis. A Big Five questionnaire asks participants to rate statements about behavior and experience, such as being quiet around strangers, paying attention to details, worrying about things or enjoying abstract ideas. Researchers care about latent causal variables such as openness, conscientiousness, extraversion, agreeableness and neuroticism, which are inferred from patterns across many responses. Causal discovery is valuable in this setting because it can organize many correlated responses into structured causal hypotheses about questionnaire statements.

The dataset contains 50 observed indicators, ten for each Big Five dimension, collected from close to 20,000 complete online responses after missing values are removed \citep{dong2024versatile}. Responses are coded on a five-point agreement scale and standardized before analysis. These choices already illustrate why a guided causal discovery workflow matters: the user must understand the meaning of each question, the scale, missingness handling, reverse-worded statements, standardization and whether demographic variables are included or excluded.

Latent variables are central to the example. A response such as ``I am the life of the party'' is a measurement, while extraversion is inferred from a broader response pattern. Standard observed-variable causal discovery methods can miss this measurement structure or produce graphs that are difficult to interpret as personality mechanisms. The semantics of discovered hidden causal variables also remain open. A hidden node may roughly match a known personality dimension, mix several question groups, capture response style or arise from the method's assumptions. This makes the example useful both for introducing causal discovery and for showing why agents should assist without replacing formal methods.

\begin{figure}[t]
    \centering
    \includegraphics[width=\linewidth]{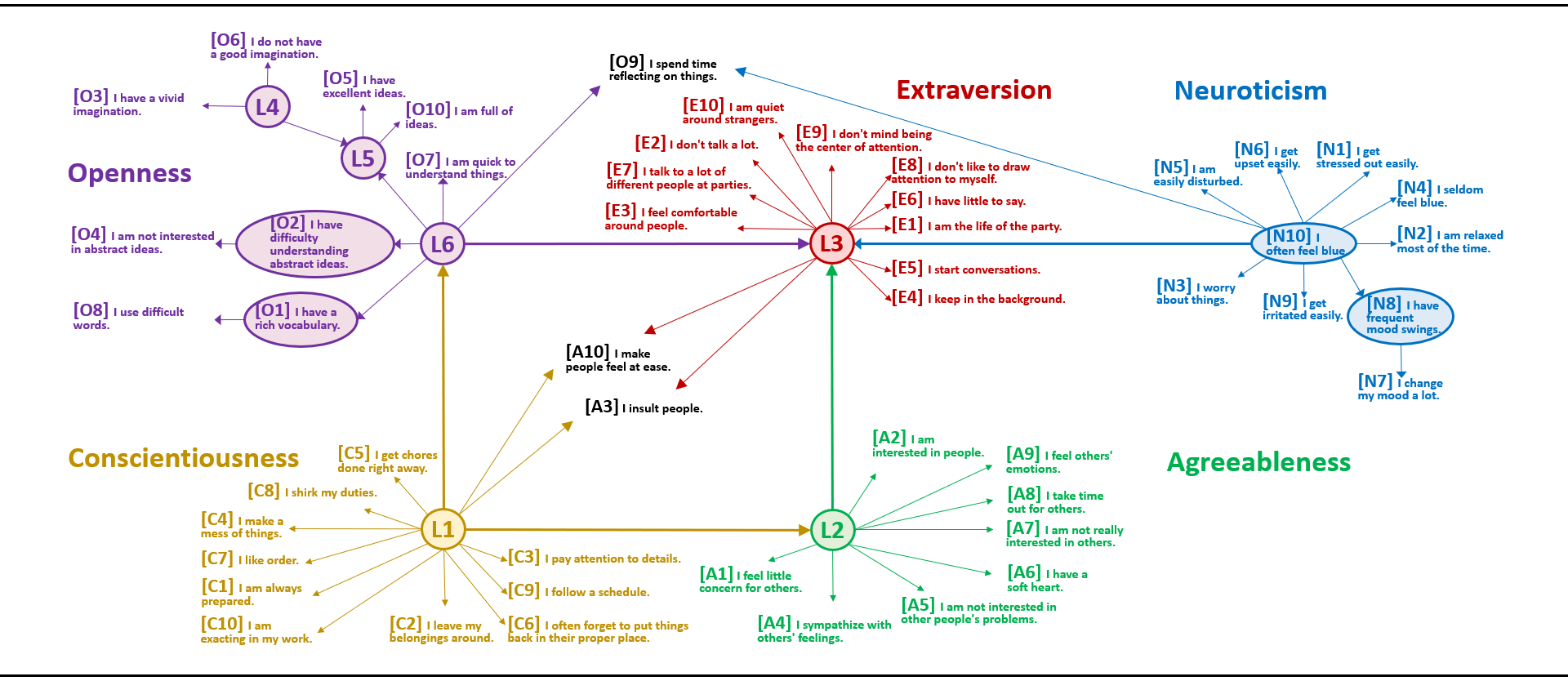}
    \caption{Causal discovery results on the Big Five personality questionnaire data. The discovered graph aligns with existing psychological knowledge and has been validated by domain experts, while also revealing several previously unexplored relationships that may warrant further investigation \citep{dong2024versatile}.}
    \label{fig:big-five-rlcd}
\end{figure}

To handle the complexity introduced by the latent variables and the flexible causal structure, the analysis uses RLCD \citep{dong2024versatile} for partially observed causal models with causally related hidden variables, together with GIN to help orient relations between hidden variables \citep{xie2020generalized}. The resulting causal graph (Figure \ref{fig:big-five-rlcd}) illustrates why causal discovery can be practically helpful. Many observed indicators associated with the same Big Five dimension are grouped near common hidden variables, consistent with the view that personality dimensions help explain response patterns. The graph also contains relations among hidden variables and between selected observed responses, closer to network views in personality research where responses may influence one another. For example, the analysis discusses latent paths such as $L1 \rightarrow L6 \rightarrow L3$ and $L1 \rightarrow L2 \rightarrow L3$, shared hidden influences on some indicators, and observed links such as $O2 \rightarrow O4$ and $O1 \rightarrow O8$ \citep{dong2024versatile}. These patterns provide hypotheses for expert inspection.

This is where agents can help. During data analysis, an agent can summarize the 50-question roster, identify the five intended dimensions and flag reverse-worded or missing responses. During preprocessing, it can explain why standardization is used and how exclusions could affect the result. During method recommendation, it can explain why a latent-variable method is appropriate. During expert-knowledge incorporation, it can retrieve background on Big Five measurement, question wording and competing latent-trait versus network interpretations. During interpretation, it can translate graph marks and latent-variable caveats into accessible language.

A careful agent response would say that the graph suggests an assumption-conditioned latent structure. It may note that a cluster resembles conscientiousness, or that $O2 \rightarrow O4$ is compatible with a hypothesis about abstract ideas, while presenting both as hypotheses for domain review. It should avoid naming hidden variables as confirmed traits, inferring psychological mechanisms from a single arrow, adding orientations, removing responses or tuning rank thresholds to match an expected Big Five story. Formal diagnostic, domain review and explicit user judgment remain necessary.

The value of \textit{causal-learn+} in this example lies in making a complex causal discovery workflow both accessible and transparent. Agents assist users in understanding the data, selecting appropriate methods, documenting preprocessing decisions, and interpreting complex outputs, while the resulting graph remains firmly grounded in the data, assumptions, discovery procedure, diagnostic analyses, and domain knowledge. More broadly, the example illustrates the complementary roles of causal discovery and agents: causal discovery provides the methodological foundation for causal conclusions, whereas agents help users effectively apply, interpret, and communicate those conclusions.

\section{Discussion}

The arrival of agents does not change what counts as causal evidence. In causal discovery, a graph mark is meaningful only through the data, assumptions, formal procedure, diagnostics and scientific decisions that produced it. Agent fluency can help people reach and understand this evidence, but fluency cannot create evidence. This distinction is the central claim of agentic causal discovery with trust.

The design standard should be simple: agents assist; algorithms discover; scientists judge. Agents can make causal discovery more accessible by helping with data understanding, preprocessing, method choice, expert context, tool use and interpretation. The inferential core should remain explicit and inspectable, with conditional-independence tests, scores, rank constraints, graph search, orientation rules and outputs tied to formal methods and user-approved assumptions. This discipline is the condition for trust.

Future work should make this standard operational. The community needs benchmarks that test whether agents improve data diagnosis, method selection and calibrated interpretation; audit trails that separate data, assumptions, agent recommendations, user decisions and algorithmic outputs; interfaces that let experts approve assumptions before they enter the analysis; and evaluations of whether agent guidance reduces overinterpretation of partially identified graphs. Agentic causal discovery should be judged by whether it lowers the barrier to rigorous causal analysis while preserving the standard of causal evidence. A more ambitious direction is to train foundation models for causal analysis from causal principles, with explicit assumptions, identifiability conditions and failure modes built into their objectives and evaluations. If such models can produce causal outputs with formal guarantees, agents may eventually participate more directly in causal discovery itself. Until then, the safe design is to keep today's agents outside the inferential core while using them to make that core easier to access, inspect and interpret. This framework is an initial step toward that goal.

\bibliography{sample}

\end{document}